# When Do Agent Loops Mistake Stagnation for Progress?

## Self-Evaluation Bias and Externally Grounded Verification in Long-Running Autonomous LLM Agent Loops

Hyundoo Park · Byungho Choi

*Independent Researchers, Republic of Korea*

**Abstract.** Long-running autonomous agents increasingly plan, act, and judge their own completion without continuous human intervention. When an agent grades its own work, however, self-evaluation bias takes hold: plausible changes are accepted as progress, and the loop reports advancement while real-world outcomes stagnate or regress. We name this failure mode the progress mirage and show, with controlled measurement, that it is a question of what the evaluator is grounded in. We built a testbed that holds the agent and its tool surface fixed across all conditions and manipulates only the information-channel type of the evaluator that gates the loop. A world-state oracle (a synthetic user simulator) that is unfakeable in principle is enforced by container and network isolation, and the agent's inability to reach that oracle is empirically verified by a boundary proof at every run and a canary scan at every cycle. Running a real frontier agent repeatedly under three evaluator arrangements yields the following results. The agent claimed its change was an improvement in all 54 cycles, yet 56 percent of those cycles had a measured delta of zero or below. Self-report was thus an uninformative signal, and the self-verdict gate that trusts it degenerated into accept-all, eroding the best deployed state it had reached by 19 percent through subsequent acceptance of regressions. Even the strongest in-band judge — one that reads the full artifact text, the change diff, and its own verdict history — accepted cycles of which 44 percent were real-world regressions and rejected 38 percent of real improvements, so the preregistered adversarial hypothesis that a strong judge closes the gap was rejected. In contrast, on a boundary task whose success specification is verifiable from the artifact itself, the same judge's mirage vanished to 0 and the gap to out-of-band collapsed within the registered threshold (H2 accepted), showing that the gap is not an artifact of the apparatus but depends on where the success signal resides. Further, in a sign-only variant that removes the oracle numbers from the feedback and returns only the acceptance verdict, real-world output remained similar to full feedback (mean 110.0 versus 113.0), decomposing at pilot scale the benefit of out-of-band evaluation as coming from the gate's grounding rather than from the information content of the feedback. The cause of the gap is not the evaluator's quality but what the evaluator is grounded in. We generalize these results into an architecture centered on an out-of-band evaluator, report the anonymized six-week, 61-cycle field observation of a single deployment that motivated it, and situate the architecture relative to platform built-in loop/goal primitives via the distinction of loop plumbing vs. the reward brain. Our main conclusion is as follows. For open-ended real-world objectives whose success signal lives in the state of the world outside the transcript, scaling up the judge is not enough; out-of-band evaluation with access to the real world is a structural requirement for preventing the progress mirage.

> **Draft note.** This manuscript is a preliminary draft comprising a controlled pilot measurement and the anonymized single-deployment field observation that motivated it. The measurement holds the agent fixed and manipulates the information-channel type of the evaluator; the field observation is presented as motivation only. Generalization to multiple models and tasks, and the full measurement after the preregistration freeze, are deferred to follow-up work.

## 1 Introduction

Over the past few years, agents built on large language models (LLMs) have moved beyond one-shot responses toward autonomously executing multi-step tasks. An agent gathers context, invokes tools, verifies results, and retries, repeating this cycle without human intervention at every step. Recently, major platforms have begun to offer this cycle as a first-class feature: a loop form that re-executes a prompt at fixed intervals, and a goal form in which, once a completion condition is defined, a separate small model judges at every turn whether it is satisfied. The focus of the work thereby shifts from "instructing the agent on individual tasks" to "designing the cycle that issues the prompts." We call this design activity loop engineering.

As cycles grow longer and human intervention becomes rarer, one question becomes decisive: who judges completion? The simplest answer is that the agent judges itself. But an evaluator that grades its own work is systematically biased toward leniency. This self-evaluation bias has been observed repeatedly in preference-model and LLM-as-judge research [Zheng et al., 2023], and it connects to reports that self-correction does not work reliably without external signals [Huang et al., 2024]. Self-reflection techniques such as Reflexion and Self-Refine depend directly on the quality of self-evaluation [Shinn et al., 2023], [Madaan et al., 2023]; when the evaluation signal is weak, iteration can converge not on improvement but on elaborate stagnation.

Elaborate stagnation is precisely the failure mode this paper addresses. If the reward function is set up to score outputs (analysis, new frames, self-checks) as progress, a sufficiently capable agent will optimize that reward. Every cycle produces new analysis and passes the verdict criterion. Progress accumulates in the loop's own logs. Yet real-world metrics — users, revenue, indexed pages — do not move. This is an instance of reward misspecification and specification gaming [Amodei et al., 2016], [Krakovna et al., 2020], [Skalse et al., 2022]. An agent optimizing a proxy reward drifts away from the true goal to the extent that the proxy diverges from it, and as autonomy grows, this divergence persists longer and with greater confidence.

The heart of the problem lies in what the evaluator can see. Platform built-in judges are generally in-band: they judge based only on what the agent has placed into the conversation within the same session. When the task is well-bounded and verifiable from the transcript alone (tests pass, the build succeeds, a file exists), this suffices. But when the goal is open-ended and grounded in the real world — when the evidence of completion resides not in the conversation but in the state of external systems — an in-band evaluator is structurally unable to see that evidence. If the agent writes "the metrics improved" into the transcript, the in-band evaluator can only assess the statement; it has no means of verifying its truth. The evaluator is trapped inside the narrative the agent constructs.

### 1.1 Our Approach

We propose an architecture that places the evaluator out-of-band. The evaluator runs as a separate process, apart from the agent, and independently accesses real-world ground truth rather than the conversation transcript. Concretely, it reads databases, version-control history, web analytics, and append-only ledgers with its own tools. The evaluator enforces a reward function that asks whether real-world metrics actually moved, and for cycles that fail this test it leaves a rejection artifact that the next loop iteration is forced to read first at boot. This closes the path by which a cycle could terminate on self-evaluation alone. Figure 1 summarizes this structure.

An out-of-band evaluator alone is not sufficient, and we present four auxiliary mechanisms around it. These auxiliary mechanisms are design notes derived from field observation; they are not individually validated by our controlled measurements. First, state is recorded only as append-only logs or views derived from them, because a field overwritten in place becomes a device that manufactures falsehood. Second, handoff between sessions happens through files, not conversation. Third, a frame-adversarial review makes the agent periodically attack its own deepest assumptions. Fourth, a self-healing scheduler revives dead sessions.

We support our claims with two kinds of evidence. First, as motivation, we present an anonymized six-week, 61-cycle field observation of a single production deployment. In this deployment, every evaluated cycle was accepted under an in-band self-verdict that rewarded analysis, while real-world outcomes stagnated; when the reward function was moved to real-world outcomes and enforced out-of-band, the behavior reversed. A single observation, however, cannot establish causality. The core evidence of this paper is therefore controlled measurement. We built loop-engineering-lab, a testbed that holds the agent and its tool surface fixed and manipulates only the information-channel type of the evaluator gating the loop; we enforced a world-state oracle that is ungameable in principle via container and network isolation; and we ran a real frontier agent repeatedly under three evaluator arrangements, measuring the progress mirage and real-world output.

The measurements are as follows. The agent claimed in every cycle that its change was an improvement, and evaluators bound to the transcript accepted as progress a substantial fraction of the cycles in which real-world metrics stagnated or regressed. The self-verdict loop eroded the best deployed state it had reached by accepting subse-quent regressions, while the out-of-band evaluation loop grounded in the world-state oracle reached higher real-world output than the self-verdict loop. Yet even the strongest in-band judge — one that reads the transcript critically — failed to close this acceptance-error gap. When the success signal lives in the state of the world outside the transcript, the cause of the gap is not the evaluator's quality but what the evaluator is grounded in. Conversely, on a boundary task where success is verifiable within the transcript, the gap narrowed sharply. This falsification control supports the claim that our results are not the product of an apparatus designed to favor a particular arrangement, but the effect of evaluator grounding.

### 1.2 Contributions

Our contributions are as follows.

1. We formalize the progress mirage — the failure mode in which a long-running autonomous agent mistakes analysis and process for progress, so that the loop reports advancement while real-world outcomes stagnate — as a measurable phenomenon, and diagnose its cause as the in-band structure that traps the evaluator inside the transcript.
2. We present loop-engineering-lab, a testbed that measures this phenomenon. For an open-ended task that cannot be verified by tests, it installs a synthetic-user world-state oracle, enforces the agent's in-principle inability to reach that oracle via container and network isolation, and empirically verifies it with a boundary proof at every run. The measurement apparatus that exposes a new failure mode is itself a contribution.
3. We measure the causal effect of evaluator grounding under control. Holding the agent fixed and swapping only the evaluator's information-channel type — in-band-self, a strong in-band-judge, and out-of-band — we measure the progress-mirage rate, real-world output, and the number of cycles to the first real outcome. This design includes a preregistered adversarial hypothesis testing whether the strongest in-band judge closes the advantage of out-of-band.
4. We generalize these results into an architecture centered on an out-of-band evaluator, and situate it relative to platform built-in loop/goal primitives via the distinction of loop plumbing vs. the reward brain. We argue that plumbing (scheduling, pacing, session revival) can be delegated to built-in primitives, whereas the reward brain grounded in the real world cannot. We also make explicit where the boundary lies within which in-band evaluation suffices.

### 1.3 Organization

Section 2 reviews related work. Section 3 describes the architecture under test and the measurement apparatus. Section 4 presents the measurement results together with the motivating field observation. Section 5 discusses implications and trade-offs. Section 6 addresses threats to validity, and Section 7 concludes.

## 2 Background and Related Work

This work concerns the loops of long-running autonomous LLM agents. Our central claim is that when an agent loop grades its own completion, self-evaluation bias causes it to accept process and analysis as progress while real-world outcomes remain stagnant. We name this phenomenon the *progress mirage* and propose an architecture that separates the generator (the agent) from the evaluator, placing the evaluator in a separate out-of-band process with independent access to ground-truth signals (databases, version control, web analytics, append-only ledgers) rather than the agent's conversation transcript. This section situates that contribution within five clusters of literature, and at the end of each cluster makes explicit the gap our work fills.

### 2.1 Autonomous LLM Agents and Self-Reflection That Depends on Self-Evaluation

The paradigm of using an LLM as an agent was established by ReAct, which interleaves reasoning and acting [Yao et al., 2023]. ReAct showed a synergy between the two capabilities: reasoning traces guide action plans, and actions collect observations from the external environment. Subsequent work extended this into a self-reflection family in which the agent critiques and improves its own outputs. Reflexion reinforces the agent with verbal feedback instead of weight updates, accumulating the agent's own reflections on prior failures in episodic memory to improve decision-making in the next attempt [Shinn et al., 2023]. Self-Refine likewise proposes an iterative refinement loop in which the same LLM plays all three roles of generation, feedback, and refinement [Madaan et al., 2023].

The shared premise of these methods is that the model can reliably judge the quality of its own outputs. Huang et al., however, showed that intrinsic self-correction without external feedback fails to improve performance on reason-ing tasks and sometimes degrades it [Huang et al., 2024]. This critical result suggests that loops relying solely on self-evaluation may fail to produce accurate progress signals. All of this literature, however, targets one-shot tasks or short bundles of attempts; it does not address how a long-running autonomous loop spanning multiple context windows distorts its self-grading relative to ground-truth outcomes.

### 2.2 LLM-as-Judge and Self-Evaluation Bias

Why self-evaluation cannot be trusted has been systematically established in the LLM-as-judge literature. Zheng et al., proposing MT-Bench and Chatbot Arena, which use strong LLMs as evaluators, reported that LLM judges exhibit position bias, verbosity bias, and self-enhancement bias [Zheng et al., 2023]. Self-enhancement bias in particular — the evaluator's tendency to prefer outputs similar to its own — is a direct threat to self-grading loops. Panickssery et al. characterized this more precisely, showing that LLM evaluators can recognize their own generations (self-recognition) and that a linear correlation exists between the strength of this recognition ability and the strength of self-preference bias [Panickssery et al., 2024].

This cluster establishes that evaluation is structurally biased when generator and evaluator are the same model. These studies, however, measure preference verdicts over static response pairs and restrict the evaluator's input to model-produced text (responses, transcripts). Recent work that widens the target to agent trajectories rests on the same limitation. AgentRewardBench quantified the systematic errors of LLM-judge verdicts on web-agent trajectories against human ground-truth labels [Lù et al., 2025], but does not treat the evaluator's information channel itself as a manipulated variable. Chain-of-thought monitoring, in which another model watches the reasoning process, likewise showed that as long as the monitoring signal remains in model-produced text, the agent can be optimized to circumvent that channel [Baker et al., 2025]. We hold that the way to escape this bias lies in separating the evaluator's *input* from text: what the evaluator grades must be not the transcript in which the agent narrates what it did, but ground-truth signals recorded in external systems.

### 2.3 Reward Hacking and Specification Gaming

The failure in which a self-grading loop mistakes process for progress is isomorphic to reward hacking and specification gaming in the reward-specification literature. Amodei et al. formalized reward hacking — arising when the objective function the designer wrote down admits "easy" solutions that distort its intent — as a core accident risk in AI safety [Amodei et al., 2016]. Krakovna et al. named behavior that satisfies the literal specification of an objective without achieving the intended outcome specification gaming, cataloguing many cases in which reinforcement-learning agents find shortcuts that collect reward without completing the task as intended [Krakovna et al., 2020]. Skalse et al. gave the first formal definition of reward hacking, defining a proxy as unhackable only when increasing the expected proxy reward can never decrease the expected true reward [Skalse et al., 2022]. Pan et al. demonstrated empirically that more capable agents exploit the gaps in misspecified proxy rewards more effectively, raising proxy reward while lowering true reward, with a phase transition at a capability threshold where true reward drops sharply [Pan et al., 2022].

This risk has also been characterized quantitatively. Gao et al. showed scaling laws for overoptimization, in which optimizing against a proxy reward model beyond a certain scale actually harms true reward [Gao et al., 2023], and Manheim and Garrabrant classified the failures that arise when a proxy stands in for a goal as variants of Goodhart's law [Manheim and Garrabrant, 2018]. This cluster provides, in general form, the core problem of our work: the divergence between proxy and true reward. Our progress mirage can be read as an LLM-agent-loop-specific instance in which completion verdicts based on self-evaluation are themselves a hackable proxy reward, and the more capable the agent, the more effectively it optimizes this proxy while real-world outcomes stagnate. The reward-hacking literature, however, mainly targets reward functions in reinforcement-learning environments and does not treat the termination condition of a long-running autonomous LLM agent loop as a reward-specification problem.

### 2.4 Generator-Evaluator Separation, RLHF Reward Modeling, and Process vs. Outcome Supervision

The design that separates the evaluator from the generator is already established in RLHF reward modeling. Christiano et al. proposed deep reinforcement learning from human preferences, learning a separate reward model from human preferences to optimize the policy, thereby structurally separating the generating policy from the evaluative reward signal [Christiano et al., 2017]. Ouyang et al.'s InstructGPT applied this separation to large-scale language-model alignment, fine-tuning the policy with a reward model trained on human feedback over output rankings [Ouyang et al., 2022]. Bai et al.'s Constitutional AI externalized the source of the evaluation signal into a set of principles, performing alignment based on AI feedback [Bai et al., 2022].

The question of where the evaluation signal is taken from also connects to the process vs. outcome supervision debate. Uesato et al. systematically contrasted process-based and outcome-based feedback in mathematical problem solving [Uesato et al., 2022], and Lightman et al. showed that process supervision, which gives feedback at every intermediate reasoning step, outperforms outcome supervision, which gives feedback only on the final result, on mathematical reasoning tasks [Lightman et al., 2023]. From our perspective, this contrast shows that the choice of evaluation target is decisive.

The scalable-oversight family, which seeks to extend supervision by strengthening the evaluator — debate, which elicits truth through argumentation [Irving et al., 2018]; recursive reward modeling, which recursively delegates evaluation itself to learned reward models [Leike et al., 2018]; and weak-to-strong generalization, in which a weak supervisor guides a strong model [Burns et al., 2023] — commonly focuses on building better judges. Our adversarial hypothesis HA1 directly tests exactly this premise — that raising judge quality makes supervision sufficient — on open-ended real-world objectives, and its rejection suggests there are domains where the ceiling is not judge quality but the judge's information channel. But the evaluation signals of RLHF reward models and process supervision all take model-produced text (responses, reasoning steps) as input. Our work inherits the generator-evaluator separation principle of reward modeling but differs in replacing the evaluator's input with ground-truth signals from external systems rather than model-produced text, and in placing the evaluator as a separate process operating inside the loop at execution time rather than at training time.

### 2.5 Autonomous Software-Engineering Agents and Harness Design

The literature on long-running coding agents and their harnesses provides the most direct precedent for grounding evaluation in ground truth. SWE-bench has models modify a codebase on tasks built from real GitHub issues and corresponding pull requests, judging completion not by the model's self-report but by running the repository's test suite [Jimenez et al., 2024]. SWE-agent designed an agent-computer interface that lets the agent edit files, navigate the repository, and run tests, raising performance substantially [Yang et al., 2024]. These studies demonstrate that execution-based grading is more reliable than self-report, but they use evaluation as a one-time grading step for a benchmark, not as a standing in-loop reward function that enforces the termination condition of a running loop.

Industry engineering guidance addresses this gap more directly. Anthropic's guidance on building effective agents presents design principles for raising the complexity of agentic systems only as much as needed and for securing reliability when agents operate autonomously across multiple turns [Anthropic, 2024]. Harness guidance for long-running agents goes a step further, recommending a structure that combines progress files with version-control history so the agent maintains progress across context windows, and that has the next session confront, as ground truth, a list marking each feature "passing" or "failing" [Anthropic, 2025]. Related harness-design guidance likewise holds that each component of a harness encodes an assumption about what the model cannot do on its own [Anthropic, 2025b]. These engineering guidelines connect directly to our concern, but they have not been formalized into a peer-reviewed architecture, and they do not generalize the approach of isolating the evaluator into an out-of-band separate process without access to the agent's transcript, defining a reward function over plural real-world signals such as databases, web analytics, and append-only ledgers.

### 2.6 Direct Comparison with Contemporaneous Measurement Studies

The exact combination we address — the causal effect of the evaluator's information-channel type (transcript reading vs. world-state querying) on self-deception under open-ended real-world objectives — is closest to a cluster of measurement studies that formed rapidly in late 2025 and 2026. We therefore contrast this cluster individually and make explicit the empty cell our work occupies.

SpecBench quantifies reward hacking in long-horizon coding agents as the pass-rate gap between visible validation tests and hidden held-out tests, showing that the gap grows steeply with task length [Zhao et al., 2026]. It shares our concern with the in-band vs. out-of-band gap, but SpecBench's ground truth is held-out unit tests and the evaluation channel on both sides is test execution. Our work differs in that (a) it takes as ground truth a world-state oracle (synthetic user conversions) that cannot be verified by tests, and (b) it treats the evaluator's informa-tion-channel type itself as the manipulated variable while holding the agent fixed.

Pan et al. showed spontaneous reward hacking in iterative self-refinement on open-ended essay tasks, where the LM evaluator's score rises while actual quality stagnates or declines [Pan et al., 2024]. This is the same phenomenon as our progress mirage, but their evaluator is another text-reading judge, and they include no world-state oracle, no channel-type manipulation, and no judge-quality ablation.

EvilGenie reports that on coding tasks an in-band LLM judge is highly effective at detecting reward hacking and that held-out tests add little [Gabor et al., 2025]. This result appears at first to conflict with our claim, but it is in fact consistent with our boundary hypothesis H2. In coding settings where the hack is identifiable (unambiguous) from the transcript, a strong in-band judge suffices. Our claim is restricted to open-ended objectives whose success signal lives in world state outside the transcript, and we measure both regimes within the same experiment to draw that boundary. EvilGenie therefore reads not as a counterexample but as external confirmation of the boundary condition.

Çağatan and Zhao show, in text-based AI Safety Gridworlds, a separation between observed reward and hidden safety objectives, reporting that standard reinforcement-learning optimization widens the gap [Çağatan and Zhao, 2026]. This is a toy RL environment, not a real-world oracle responding to deployed artifacts. RewardHackingAgents benchmarks evaluation-integrity attacks in which ML-engineering agents tamper with metric computation or induce train/test leakage, together with defenses (evaluator locking) [Atinafu and Cohen, 2026]. That is a threat model in which the agent actively attacks the evaluator, whereas we isolate the agent by network so that it cannot reach the evaluator (the oracle) at all and empirically verify this unreachability at every run; the threat models differ and the two works are complementary.

Two further adjacent strands exist. First, the line that evaluates agents with synthetic users. tau-bench grades an agent by comparing the final state of a database after conversation with a simulated user against a ground-truth answer [Yao et al., 2024], sharing raw material with our oracle (world state produced by a synthetic user simulator). But tau-bench's grading is one-time grading on a verifiable benchmark with a specified answer state, and it does not treat the evaluator arrangement as a manipulated variable. Second, a line measuring the divergence between agents' completion claims and actual success formed in the first half of 2026. Advani characterized false success — agents confidently declaring completion while the environment state is failure — on large trajectory sets from verifiable benchmarks (tau2-bench, AppWorld), showing that no combination of five judges and five prompts detects it reliably [Advani, 2026]. That result is directionally consistent with our rejection of HA1, with two differences: Advani's tasks are verifiable, with programmatic ground truth, and the judges are merely compared against fixed ground truth — the evaluator's information-channel type itself is not manipulated. Wang et al. compared four verifier configurations for coding-agent reward — tests, rubrics, user verification, and agent verification — showing that none is a panacea [Wang et al., 2026], and Flynt, starting from cases where frontier judges score ungrounded responses highly, proposes deterministic trajectory-based grading [Flynt, 2026]. In the context of agent-organization design, Xie reported a large divergence between self-reported success rates and measured recovery rates as agent self-deception [Xie, 2026]. All of these confirm, from different angles, the existence of the phenomenon we target, but none performs a controlled measurement that manipulates only the evaluator's information-channel type on non-verifiable open-ended tasks whose success signal is in principle outside the transcript.

New adjacent work identified in a pre-submission rescan (2026-07-23) also falls within the same map. ScopeJudge performs an ablation over five levels of transcript context visible to a judge that gates a security agent's tool calls before execution [Caldwell et al., 2026]. The grammar of manipulating what the evaluator sees is shared with our design, but all five levels vary the amount of in-transcript context, and the object of judgment is compliance with a pre-declared scope rather than progress on an open-ended goal; a world-state channel outside the transcript appears at no level. On the training side, RePro trains agents to self-generate progress signals [Ma et al., 2026]; since this strengthens in-band self-signals, our results supply the boundary condition for where such signals can be sound (tasks whose success signal is verifiable inside the transcript). Work on early diagnosis of wasted computation in multi-agent systems shares our concern with wasted cycles but does not manipulate the evaluator's information channel [Li et al., 2026].

In sum, the uniqueness of this work lies in the combination of three anchors, with the center of gravity on the second. (1) We ground a non-verifiable open-ended task, for which no answer specification exists in principle, with a world-state oracle; (2) with the same agent and the same task, we manipulate only the information-channel type of the evaluator gating the loop (to our knowledge this design appears nowhere in the cluster above); and (3) we separate the irreducibility of grounding vs. judge quality with an ablation inside the same apparatus. Since anchor (3) alone already receives strong evidence from Advani's results, we position (3) as an auxiliary pillar to (2). Given how fast the field is moving, we state this combination as our criterion of occupancy and preregister the position that, should work performing the identical combination appear, we will narrow our claim to the residual gap.

### 2.7 Summary of the Research Gap

In sum: the self-reflection literature (2.1) builds loops that depend on self-evaluation but assumes its reliability; the LLM-as-judge literature (2.2) establishes self-evaluation bias but restricts the evaluation input to model-produced text; the reward-hacking literature (2.3) generalizes the proxy vs. true reward divergence but does not apply it to the termination conditions of LLM agent loops. The reward-modeling and process-supervision literature (2.4) establishes generator-evaluator separation but takes the evaluation signal from text and places it at training time. The autonomous software-engineering agent and harness literature (2.5) demonstrates execution-based ground-truth grading but does not formalize it as a standing reward function of a running loop.

Our work fills the following gaps. First, we formalize the progress mirage — self-evaluation bias mistaking process and analysis for progress in long-running autonomous LLM agent loops — as a reward-specification problem. Second, we separate the generator (agent) from the evaluator, isolating the evaluator into an out-of-band separate process without access to the agent's conversation transcript, and define a real-world reward function over plural ground-truth signals: databases, version control, web analytics, and append-only ledgers. Third, the evaluator records rejection artifacts that the next loop iteration is forced to confront, so that the loop's termination condition is governed not by in-band self-verdicts but by out-of-band evaluation anchored in real-world signals. These three are the loop-engineering contributions this paper offers in place of transcript-based in-band self-judgment.

## 3 Method: The Architecture Under Test and the Measurement Apparatus

This section consists of two parts. The first half describes the architecture we propose and validate — a design centered on an out-of-band evaluator. This architecture is the treatment manipulated in the measurement. The second half describes loop-engineering-lab, the apparatus that measures the treatment's causal effect under control. The core of the measurement is to hold the agent and its tool surface fixed across all conditions and to swap only the information-channel type of the evaluator that terminates the loop. We first cover the design principles, the out-of-band evaluator, the real-world reward function, the four auxiliary mechanisms, the cycle protocol, and the three layers of cadence, and then describe in turn the apparatus's open-ended task, the world-state oracle, isolation and boundary proof, the three evaluator arms, the dependent variables, and the pilot scope.

### 3.1 Design Principles

The architecture follows three principles. First, the entity that works and the entity that judges are separated. Second, the judging entity looks at the state of the world, not at the narrative constructed by the working entity. Third, the outcome of the verdict is recorded in a form the next iteration cannot evade. These three principles are realized, respectively, as generator-evaluator separation, out-of-band access, and forced feedback.

### 3.2 Generator-Evaluator Separation and the Out-of-Band Evaluator

The generator is the agent. The generator doubts, forms hypotheses, sets direction, delegates work to a worker that executes it, and reviews the results in the first instance. But because the generator's self-review is exposed to self-evaluation bias, it holds no authority to close a cycle. That authority belongs to the evaluator.

The evaluator does not operate inside the same session as the generator. It runs as a separately scheduled process and accesses real-world ground truth directly, rather than the generator's conversation transcript. Concretely, the evaluator reads the operational database, version-control history, web analytics, and an append-only ledger with its own tools. This separation is the crux. An in-band evaluator has no means to verify the truth of the statement when the agent writes "the metrics improved," whereas an out-of-band evaluator puts the same question directly to the world. Figure 1 summarizes this structure.

This design belongs to the same lineage as generator-discriminator separation and the reward-model separation of RLHF [Christiano et al. 2017], [Ouyang et al. 2022], but is distinguished by placing the evaluator's input in external system state rather than in the model's internal representations or the conversation.

### 3.3 The Real-World Reward Function

The reward function the evaluator enforces asks about outcomes, not process. That is, it asks not "did this cycle produce new analysis" but "did real-world metrics actually move." As the reward-misspecification literature has shown, an agent optimizing a proxy reward drifts from the goal to the extent that the proxy diverges from it [Pan et al. 2022], [Skalse et al. 2022]. The reward function must therefore be tied to a signal as close to the goal as possible, not to a proxy.

Concretely, the reward function forces every cycle to read the real-world arrival ledger first and compute how many days the metrics have been flat. When stagnation exceeds a threshold, an operating mode is forcibly switched on, and in that mode ACCEPT is possible through only three paths: deploying a real acquisition mechanism live, deepening product value verified against external (non-internal) personas, or escalating the binary forcing function upward. Cycles that only produce analysis or only touch policy wording are blocked in this mode. This transposes the distinction between process reward and outcome reward into an operating rule [Lightman et al. 2023].

### 3.4 Rejection Artifacts and Forced Feedback

The evaluator's verdict does not evaporate. If the real-world signal is zero and the cycle presents no valid evidence (e.g., a version-control commit or a live URL), the evaluator leaves a rejection artifact as a file. The next loop iteration is wired to read this rejection artifact first at boot. This closes the path by which a cycle could terminate on self-evaluation alone, and the previous cycle's failure becomes the next cycle's first input. Because the evaluator is separated from the generator and looks at the world directly, even if the agent dresses up stagnation as progress, it will face that dressing-up at the next boot.

### 3.5 Append-Only State

State is recorded only as append-only logs or views derived from them. Overwriting the same field in place is forbidden. If in-place updates are allowed, the state file becomes not a record of truth over time but a device for re-narrating the past to fit the current story — a channel through which self-evaluation bias seeps into state itself. When a derived view is needed, it is computed deterministically from the append-only log.

### 3.6 File-Based Handoff

Handoff between sessions happens through files, not conversation. A long-running loop necessarily spans multiple sessions due to context limits, and if progress is carried across session changes only as conversation summaries, loss and distortion accumulate. If the artifacts, ledgers, derived views, and next-step plans are all kept in files, each subsequent session starts from the same ground truth. The importance of progress files and pass/fail criterion lists in long-running agent harness design has also been emphasized in industry reports [Anthropic 2025].

### 3.7 Frame-Adversarial Review

The agent is made to periodically and adversarially attack its own deepest assumptions. A small number of assumptions underpinning the operation are made explicit in a registry, and periodically — and whenever specific conditions are met (e.g., metrics worsening consecutively) — the agent is forced to output, for each assumption, "if this assumption is wrong, what is the alternative, and what are the grounds for keeping or discarding it." Given the observation that self-reflection techniques can converge on stagnation without external signals [Huang et al. 2024], this adversarial ritual functions as a device for surfacing frame-level errors that self-evaluation misses. The efficacy of this mechanism is, however, capped by model capability; we discuss this limit in Section 6.

### 3.8 Self-Healing Scheduler

Because the loop must operate for long periods without continuous human intervention, recovery from missed boots and session deaths is required. A separate scheduler checks the elapsed time of the last active session and reboots sessions that have stalled past their normal period, restricted to active hours. This mechanism belongs to operational plumbing, not to the essence of the architecture, and can be replaced by platform-provided scheduling primitives. We return to this distinction in Section 5.

### 3.9 Cycle Protocol

A cycle proceeds in the following order. Figure 2 summarizes it.

1. BOOT. Read first the rejection artifact left by the previous cycle.
2. AUDIT. Measure actual state first, not cached summaries.

3. INTENT. Pose one broad question, not a narrow classification.
4. Generate-execute ping-pong. Early phases pass through doubt and reframing, middle phases through hypothesis injection, late phases through strategic guidance; when the worker brings back results, the generator receives and reviews them again.
5. ACCEPT. The termination condition is not a cycle count but the production of a new frame (frame-transcendence).
6. STATE WRITE. Record state append-only.
7. External verification. The evaluator, in a separate process, checks real-world signals directly and leaves a rejection artifact if needed.

### 3.10 Three Layers of Cadence

Three layers of cadence overlay the cycle above. Layer A is rare human intervention injecting direction in ad-hoc sessions. Layer B is the operating headquarters' own loop, consisting of briefings, critiques, and retrospectives at fixed times. Layer C is the ping-pong loop that forcibly separates generator and evaluator. The human role is confined to Layer A's rare direction injection and permanent escalations; the loop runs the rest by itself.

### 3.11 Measurement Apparatus: loop-engineering-lab

The core claim of the architecture above is that placing the evaluator out-of-band reduces self-deception and increases real-world output. To measure this under control, we build a testbed that holds the agent and its tool surface identical across all conditions and swaps only the evaluator that gates the loop. The evaluator's information-channel type (transcript reading vs. world-state querying) is the sole independent variable, and agent capability is controlled. Any difference between conditions therefore stems solely from what the evaluator can see, not from the agent being smarter.

### 3.12 The Open-Ended Task and the World-State Oracle

The task is to edit a signup-funnel web app (landing page, funnel step definitions, marketing copy) to maximize actual signups. This task is open-ended: there are many routes to raising conversion (a clear CTA, fewer steps, copy length, lighter pages, and so on) and no single unit test that verifies a correct answer. The crux is that the success signal lies outside the agent's transcript.

Success is measured by a synthetic user simulator. Each cycle, the simulator extracts the actual properties of the deployed artifact — the presence of a CTA on the rendered landing page, page payload size, the number of funnel steps, and copy length — computes a weighted logistic conversion probability, and produces a conversion count over a fixed cohort. Within a repetition, the same cohort random numbers are used across cycles (common random numbers), so the conversion count is a monotone deterministic function of artifact features and the sign of a delta cannot be flipped by sampling noise. This conversion count is recorded in an oracle store the agent cannot access. These conversions respond only to the physical properties of the deployed artifact, not to the agent's narrative. The agent writing "I raised conversion" and the simulator's conversion count actually rising are entirely separate things.

The oracle's ungameability is guaranteed by three invariants. The values the oracle reads are (a) outside the agent's write access, (b) produced by an independent process (the simulator) rather than by the agent's statements, and (c) in principle unverifiable from the transcript alone. Because this task is grounded by a world-state oracle rather than hidden unit tests, it is distinct from the held-out tests of the coding-benchmark lineage.

### 3.13 Isolation and Boundary Proof

To make ungameability a matter of enforcement rather than convention, the two trust zones are physically separated. The agent container and the oracle container are placed on two separate networks with no route between them. The agent uses the external internet for model API calls but has no path to the oracle network at all, so even with a shell and network tools it cannot reach the oracle. The only shared surface is the agent's deployed-artifact volume, which the oracle reads read-only. The oracle store lives inside the oracle container and on no shared volume. The trusted party is the host runner, which accesses both sides for instrumentation only; only the agent — the party with the incentive to game — is isolated.

We prove this isolation by measurement, not assumption. The oracle serves a secret canary string on a port on its own network. At the start of every measurement run (each combination of arm and repetition), the trusted host fetches the canary (proving the server is alive), the agent fails to fetch it even when given the oracle's exact IP (no route), and the agent is confirmed unable to read the oracle store from its own filesystem; all of this is recorded in a per-run boundary log. Isolation is judged established only when the host obtains the canary and the agent obtains it neither by network nor by filesystem; if it is not established, measurement does not begin. In addition, every cycle the full agent-side output and artifact files are scanned for the canary string, and the apparatus is wired to abort measurement if it ever appears (zero occurrences in this measurement). All measurements in this paper were collected only with this boundary proof established.

### 3.14 The Three Evaluator Arms

The evaluator arrangement is the key independent variable; three implementations sit behind the same interface.

1. in-band-self: judges from the agent's self-report alone. In implementation, if the agent actually edits files and reports the rationale for the change, the cycle is accepted. In our measurement the agent claimed in every cycle that its change was an improvement, so this arm effectively operates as an accept-all baseline. This is not a defect; it is what this arm is designed to measure. Self-claims carry no information, and a self-verdict gate passes that uninformative signal straight through. The feedback merely returns the agent's own claims, with no ground truth.
2. in-band-judge: a separate, stronger model reads the agent's change summary, the full text of the produced files, the diff of the change against the deployed state, and its own past verdict history — but not the oracle — and judges whether the change is an improvement. The input cap is set generously and cap-hit status is logged every cycle (zero hits in this measurement). The verdict prompt is a critical prompt that demands evidence and defaults to rejection, and it even includes a domain warning that plausible decorative edits (adding social proof, lengthening copy) can bloat the page and hurt conversion (the full prompt is in the public repository). It is thus the strongest in-band baseline constructible in a single call, hinted at the oracle's penalty direction. This arm tests the adversarial hypothesis HA1.
3. out-of-band: accepts only when the oracle's actual conversion delta is positive. The feedback is an acceptance or rejection rationale containing the post-gate oracle scalars (the delta and the current conversion count). This treatment is therefore not a pure gate arrangement but the combination of a gate and ground-truth feedback — a deliberate choice that models exactly what a real deployment's out-of-band evaluator returns (threats in Section 6). To decompose this combination we include a sign-only variant that removes the oracle numbers from the feedback and returns only the acceptance verdict; its measurement is reported in Section 4.6.

Across the three arms the agent (model, tools, task, cycle budget) is identical. Only the gate evaluator and the feedback it returns to the loop differ. Each arm's trajectory is an independent rollout, so the three evaluators do not judge the same candidate sequence; comparisons between arms are made at the level of the distribution over repetitions. In addition, an independent monitor with no role in gating records the oracle's actual trajectory in all arms. This monitor is what allows the in-band arms' self-deception to be measured against an external standard.

### 3.15 Cycle Protocol and Dependent Variables

Each cycle proceeds in four steps. (1) The agent produces a candidate change. (2) The gate evaluator issues a verdict according to its arm. (3) The monitor records the oracle delta (candidate vs. deployed state). (4) On acceptance the candidate is promoted to the deployed state; on rejection the artifact is rolled back to the last accepted state and the rationale is fed back as input to the next cycle. This repeats up to the cycle budget. This gated-deployment model, in which rejected candidates are never promoted to deployment, is the device that makes the gate actually govern state. As its algebraic consequence, the out-of-band arm's deployed state is structurally monotone non-decreasing (it accepts only positive deltas), and the accept-all arm's deployed state equals the last candidate. The shape of a deployment trajectory is therefore partly a theorem of the gate rule, and Section 4.5 interprets it on that premise.

Dependent variables are computed mechanically from raw cycle logs with no human discretion. The progress-mirage rate is the fraction of evaluator-accepted cycles whose monitor-measured oracle delta is zero or below. The preregistration draft's wording was "cycles whose delta is zero," but that definition would exclude from the mirage the worse error of accepting regressions (negative deltas), so this paper uses the zero-or-below definition as primary, discloses this definitional change as an amendment, and reports numbers under both definitions (Section 4.2). real-outcome-at-budget is the conversion count of the deployed (last-accepted) state at the end of the budget. time-to-first-real-outcome is the number of cycles until the oracle delta first turns positive, with right-censoring recorded. The wasted-cycle ratio is the fraction of cycles whose oracle delta is zero or below. In addition, to view the gate's discrimination independently of base rates, we report conditional acceptance rates — the acceptance rate given delta zero-or-below and the acceptance rate given positive delta — and their complement, the false-rejection rate.

### 3.16 Pilot Scope and Preregistration Status

The measurements reported in this paper are a pilot. They comprise the open-ended task family T1 and the boundary contrast task B1 (Section 4.4), the three evaluator arms, one fixed frontier agent model (the in-band-judge's judge is a stronger model), 3 independent repetitions per arm, and a cycle budget of 6, plus a sign-only variant decomposing the out-of-band feedback, with 3 repetitions (Section 4.6). Analysis is performed by scripts using only the standard library, producing the dependent variables and the hypothesis verdict table from append-only cycle logs.

We state the preregistration's status precisely. The hypotheses and verdict thresholds were registered before measurement in documents and version-control commits, but the commit-hash freeze procedure that the registration document itself prescribed was not completed by the time of the pilot. This pilot is therefore an exploratory measurement under a pre-freeze registration draft, and both the difference between the registered definition and the paper's definition (the mirage definition above) and mid-run apparatus modifications are disclosed as amendments (the preregistration document and apparatus history in the appendix). Among the registered factors, the reward-function axis (process vs. outcome) and the hypotheses attached to it — H3 and the cost hypothesis HA2 — were not run in this pilot and no results are reported for them. Multiple models, a task with a server-side event log as a second oracle, the reward-function axis, and the full measurement after the commit-hash freeze are future work.

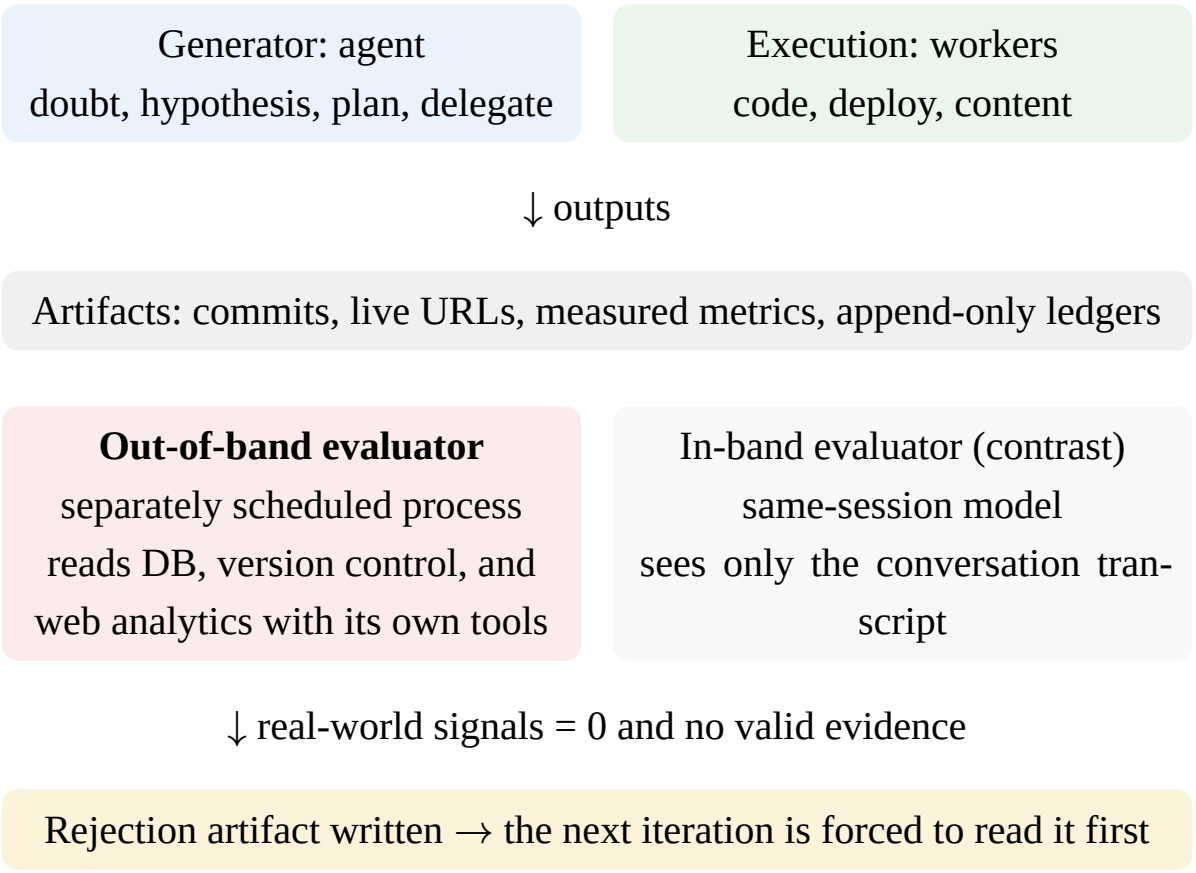

Figure 1. The out-of-band verification architecture. The generator (agent) and the evaluator are separated, and the evaluator is a separate process with independent access to real-world ground truth (databases, version control, web analytics, ledgers) rather than the conversation transcript. When real-world signals do not move, it leaves a rejection artifact that the next loop iteration is forced to confront. An in-band evaluator sees only the conversation, so verdicts grounded in real-world reward are structurally impossible for it.

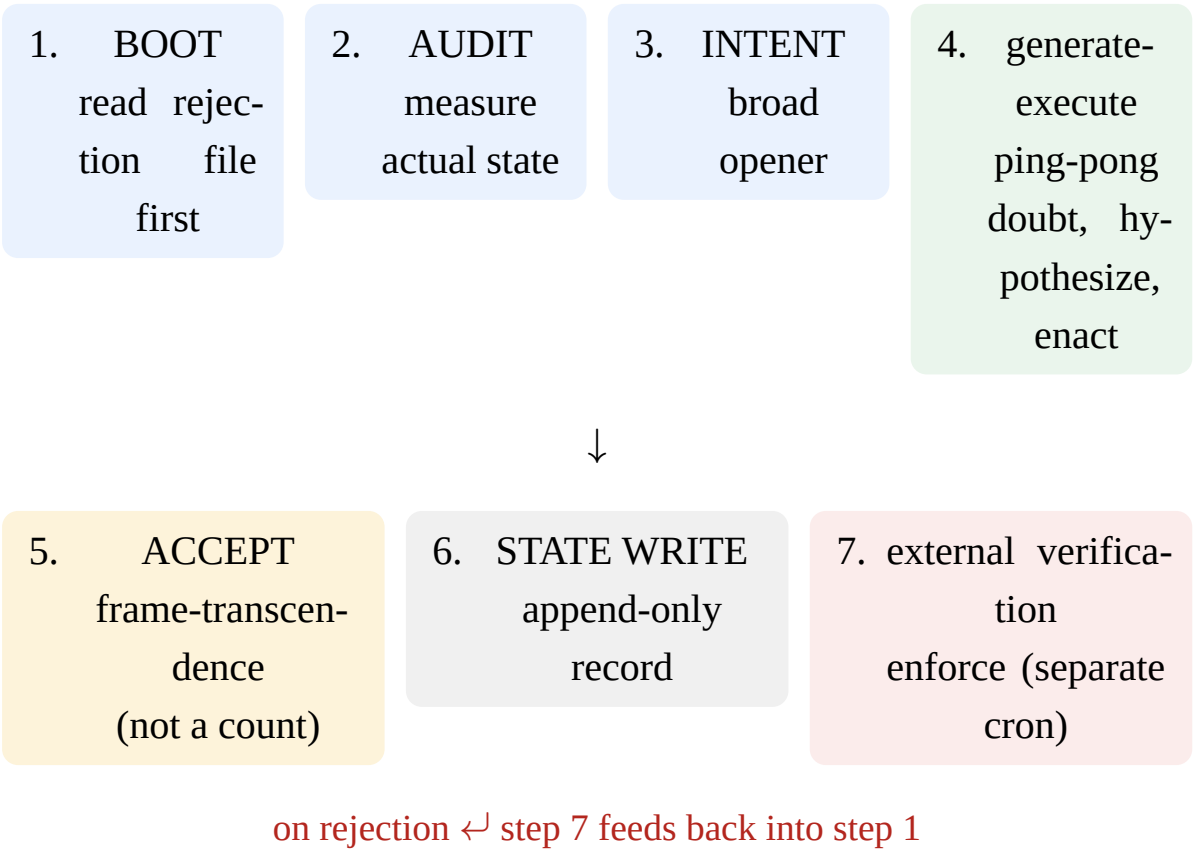

Figure 2. The loop cycle. One iteration runs from BOOT (confronting the previous rejection) through external verification, and the termination condition is frame-transcendence (production of a new frame), not a cycle count. A rejection from external verification feeds back into the next BOOT, preventing cycles from terminating on self-evaluation alone.

## 4 Measurement Results

This section reports the results of the controlled pilot collected with the apparatus of Section 3. The agent (model, tools, task, cycle budget) was held identical across all conditions, and only the information-channel type of the evaluator gating the loop was swapped among in-band-self, in-band-judge, and out-of-band. At the start of each of the nine T1 runs and nine B1 runs, the isolation boundary proof was performed and established (per-run boundary logs), and the canary scan of agent-side output and artifacts found zero occurrences in every cycle. The judge input cap was hit zero times. Analysis was performed with standard-library scripts over raw cycle logs, with no human discretion. The oracle trajectories used to judge mirage in the two in-band arms are the measurements of an independent monitor that those evaluators could not observe; in the out-of-band arm, the gate itself queries the same oracle by definition.

### 4.1 Summary of Results

Table 1 summarizes the results for the three evaluator arms. Each arm was run in 3 repetitions, the cycle budget was 6, and the mean real-world conversion of the starting (baseline) state was 33.3.

| Evaluator arm | mirage rate | (range) | Accepted cycles | P(accept\|delta <= 0) | P(accept\|delta > 0) | Deployed peak | Final deployed | Final real gain |
|---|---|---|---|---|---|---|---|---|
| in-band-self | 0.56 | (0.50, 0.67) | 6.0 / 6 | 1.00 | 1.00 | 95.0 | 76.7 | +43.3 |
| in-band-judge | 0.44 | (0.33, 0.67) | 3.0 / 6 | 0.40 | 0.63 | 100.0 | 86.7 | +53.3 |
| out-of-band | 0.00 | (structural) | 2.7 / 6 | 0.00 | 1.00 | 113.0 | 113.0 | +79.7 |

*Table 1. Measurement results by evaluator arrangement (mean over 3 repetitions per arm, budget 6). The progress-mirage rate is the fraction of evaluator-accepted cycles whose oracle-measured delta is zero or below (the primary definition of amendment A1). Parentheses are bootstrap percentile intervals over 3 trajectories and, at this sample size, should be read as observed ranges (Section 6.5). Conditional acceptance rates show the gate's discrimination independently of base rates. The out-of-band arm's mirage of 0 and deployment monotonicity (final = peak) are definitional consequences of the gate rule (Section 3). Figure 4 visualizes the mirage rates and deployment trajectories.*

The mechanical verdicts on the preregistered hypotheses are as follows. H1 is accepted: the mirage rate of in-band-self exceeds out-of-band by 0.56, far above the preregistered threshold of 0.20. The adversarial hypothesis HA1 is rejected: even the strongest in-band-judge — given untruncated input, the diff, and its verdict history — has a mirage rate 0.44 above out-of-band, the exact opposite of HA1′s prediction that the gap would close to within the 0.05 threshold. Recomputed under the registration draft's literal definition (delta exactly zero), all three arms have mirage 0.00 and H1 becomes undecidable — because that definition excludes from the mirage the worse error of accepting regressions. In this pilot, the unproductive cycles that the in-band arms accepted all had negative deltas (actual damage). We report numbers under both definitions; the rationale for the definitional amendment is in amendment A1.

The remaining registered dependent variables are also reported in full. time-to-first-real-outcome was identically 1 cycle in all arms and all repetitions (zero right-censoring). The starting state contained large first levers — introducing a clear CTA and cutting funnel steps — so the agents in all arms produced a real improvement in the first cycle; this variable therefore did not separate the arms under this task configuration (H3 was not run, along with the reward-function axis; amendment A9). The wasted-cycle ratio (fraction of all cycles with delta zero or below) was coincidentally identical at 0.56 across the three arms, meaning the regression rate of candidate generation was similar; what separated the arms was the gate's discrimination in filtering out those regression candidates. reward-hack incidence was not implemented in the pilot analysis and is not reported (amendment A4).

### 4.2 Self-Report Is an Uninformative Signal: in-band-self Degenerates into Accept-All (H1)

We begin with the most basic observation. The agent claimed in all 54 T1 cycles (100 percent) that its change would raise conversion. Yet by oracle measurement, 56 percent of those cycles had a delta of zero or below. The frontier agent's self-claimed improvement was thus uninformative as a signal of real-world progress. As its direct consequence, the in-band-self gate that trusts self-claims degenerated into an accept-all baseline, accepting all 18 cycles, and its mirage rate of 0.56 simply equals the agent's regression-generation rate. This is not a defect of the gate; it is what this arm is designed to measure. The self-evaluation channel contains no information to filter.

An honest point must be made here. The out-of-band mirage rate of 0 is not an empirical finding but a consequence of definition. Out-of-band accepts only when the oracle delta is positive, so accepting a cycle with delta zero-or-below is structurally impossible and the mirage is always 0. This value should not be read as the core evidence of this work. Our empirical claims come from two places. First, how many regressions the transcript-bound evaluators (the self-verdict, and the strongest judge of Section 4.3) actually accept as progress, and how many real improvements they reject. These values are not fixed by definition and must be measured. Second, whether that gap narrows on a task whose success signal is inside the transcript (the falsification control of Section 4.4). Together these show that our results are not merely the product of an apparatus designed to favor out-of-band.

### 4.3 Even the Strongest In-Band Judge Fails to Close the Gap (HA1)

Our adversarial hypothesis was that the weakness of in-band lies not in the information channel but in judge quality. To test it, the in-band-judge arm gave a model stronger than the agent everything: the change summary, the full text of produced files (no truncation; zero cap hits), the diff against the deployed state, and its own past verdict history — withholding only the oracle — under a critical prompt that demands evidence, defaults to rejection, and warns of the risk of decorative edits. It is the strongest in-band baseline constructible in a single call.

This strong judge was distinctly more critical than the self-verdict. It cut acceptances from 6 cycles to 3, and in conditional acceptance terms accepted only 40 percent of regressions (the self-verdict accepted 100 percent), so the direction of its discrimination was right. But the gap remained. Of the cycles it accepted, 44 percent were still real-world regressions — a gap of 0.44 to out-of-band, nine times the preregistered threshold of 0.05. At the same time, the judge rejected 37.5 percent of real improvements (false rejection). The strong judge's errors are thus bidirectional: not a problem of leniency but a ceiling on discrimination.

The judge's actual verdict rationales reveal what this ceiling is. In one cycle the agent added a styled hero section and benefit bullets; the judge accepted it as a clean, friction-reducing change, but actual conversion fell by 23 (page bloat). In another cycle it accepted concise social proof right above the CTA as adding no friction or bloat, but conversion fell by 10. In the opposite direction, it rejected the removal of a verification step (measured +7) as merely inflating unverified signups, and rejected rewrites it judged, from circumstantial evidence, to have broken links (measured +15, +16). These rationales are one and all plausible as conversion-optimization orthodoxy, and some are even sound from a real-world perspective. The problem is that nothing in the transcript says what sign that orthodoxy actually produces in this world (the simulator's response function). From the full file text and the diff, the judge could compute feature values themselves, such as page size and copy length, but it could not know which features move conversion and by how much (the sign and magnitude of the weights). The cause of the gap is not the judge's critical capacity but what the judge is grounded in. This is the most important result of our measurement; it is consistent with contemporaneous work reporting judges' failure to use information on verifiable tasks [Advani, 2026], and goes beyond it by manipulating the information channel itself to isolate the cause.

### 4.4 Falsification Control: On a Task Where Success Is Visible, the Gap Disappears (H2)

To show that our results are not merely the product of an apparatus designed to favor out-of-band, we must also measure the condition under which our claim ought to fail. That condition is the boundary task B1. In B1, success is a structural specification verifiable from the artifact itself: the number of five requirements satisfied — form presence, an email input, a submit control, exactly 3 canonical steps, and copy of 40 to 80 words. This specification is given identically to all evaluators and is verifiable from the artifact alone. A transcript-reading judge therefore holds effectively the same information as the oracle. If our claim is right, the gap between in-band and out-of-band must narrow on this task.

The results support that prediction exactly as registered. Table 2 contrasts the mirage rates on T1 and B1. The same strengthened judge's mirage rate was 0.44 on the hidden-metric task T1 but vanished to 0.00 on the visible-specification task B1. In conditional acceptance terms, the B1 judge showed perfect discrimination, rejecting every unproductive cycle and accepting every satisfying cycle, and the gap to out-of-band was 0.00, meeting the preregistered threshold (0.05 or below), so H2 is accepted. The augmentation edits the agent kept producing after the specification was fully satisfied — wording polish, adding placeholders — which do not change the satisfied count, the judge correctly judged unproductive against the specification and rejected. Set beside the fact that the same judge accepted 40 percent of regressions on T1, the conclusion is sharp: the judge's failure is not a matter of capability but of where the success signal resides, and the moment ground truth enters the transcript, the failure disappears.

One more contrast: in-band-self remained at mirage 0.50 even on the visible task. The self-verdict passes its own claims through without even checking them against the specification (accept-all), so it fails even when the success signal is in plain view. The boundary condition of H2 is thus a statement about competent in-band evaluation; the self-verdict is uninformative as a gate regardless of task type. This asymmetry — the self-verdict fails everywhere, the judge succeeds exactly when the signal is

visible — is itself evidence of falsifiability: the apparatus was not designed to make in-band lose unconditionally.

| Evaluator arm | T1 mirage (hidden metric) | B1 mirage (visible specification) |
|---|---|---|
| in-band-self | 0.56 | 0.50 |
| in-band-judge | 0.44 | 0.00 |
| out-of-band | 0.00 (structural) | 0.00 (structural) |

*Table 2. Mirage rates by task type (mean over 3 repetitions per arm, budget 6 for both tasks). T1's mirage consists mostly of accepting actual damage (negative deltas), whereas B1's unproductive cycles are delta-0 augmentation edits, so the two columns differ in kind. Out-of-band is 0 by definition on both tasks. In the first run (apparatus v1, with judge-input truncation), the B1 judge mirage was 0.17; after removing truncation and providing the diff and verdict history in v2, it dropped to 0. The v1 residual was thus an instrumentation defect (conflating state with delta), not the information channel — a fact that is itself consistent with our claim that the gap is caused by information access.*

### 4.5 Real-World Output: The Gate's Discrimination Governs the Deployed State

All three arms ended with positive final real gains (Table 1), because the starting state contained a large lever that agents in every arm captured in the first cycle. What separated the arms came after.

in-band-self accepted all six cycles and still ended lowest (from a mean peak of 95.0 down to a final 76.7, a 19 percent retreat). The erosion — a good state, once reached, being whittled away by subsequent plausible regressions — reproduced in all three repetitions. in-band-judge filtered out many large regressions, reducing erosion (peak 100.0, final 86.7), but still retreated through the regressions it accepted, and forfeited the gains of the real improvements it rejected (+7, +15, +16, and others). out-of-band ended equal to its peak in all three repetitions (113.0). As stated in Section 3, however, this monotonicity is itself a definitional consequence of the gate rule, so the empirical content here is not the monotonicity but the difference in heights reached by three arms fed candidate streams with the same regression rate (0.56). In an early apparatus variant without the gate (apparatus history in the appendix; a configuration where rejected cycles also remain deployed), final output did not separate the arms — meaning final output reveals the evaluator-channel effect only in combination with the gate rule, which is also why we treat the mirage rate and conditional acceptance rates as the primary diagnostics.

In this pilot out-of-band was also highest in final output, but we do not claim this ordering is stable from a 3-repetition pilot. What can be claimed is structural: only a discriminating gate preserves the best state reached (preventing erosion), and a transcript-bound gate cannot provide that guarantee, erring in both directions — accepting regressions and rejecting improvements.

Finally, we disclose a case that exposes the gap between the oracle and real-world value. When the agent removed the email verification step, the oracle (which counts only conversions) rewarded it as an improvement, while the judge rejected it as inflation of unverified signups. An out-of-band gate bound to a narrow metric passes such changes. This is a construct-validity problem of the oracle metric itself, separate from the in-band vs. out-of-band gap, and is discussed in Section 6.

### 4.6 Decomposing the Gate Effect from the Information Effect: Sign-Only Feedback

As stated in Section 3, the out-of-band treatment is not a pure gate placement but a combination of the gate and ground-truth feedback. The effect of the accept-or-reject verdict itself overlaps with an information effect, in which the oracle scalars carried in the feedback (the delta and the current conversion count) steer the agent's subsequent candidate generation. To decompose this combination, we additionally ran a sign-only variant on T1 with 3 repetitions, removing the oracle numbers from the feedback and returning only the acceptance verdict. The apparatus, agent model, task, cycle budget, per-run boundary proofs, and per-cycle canary scans are identical to the measurement of Section 4.1 (boundary proofs held in all 3 runs; zero canary appearances); the only thing that changes is a single feedback string. The full variant returns the oracle delta and current value along with the verdict; the sign-only variant returns the fact of the verdict alone.

| out-of-band variant | final deployed value (per repetition) | mean | accepted cycles | wasted-cycle ratio |
|---|---|---|---|---|
| full feedback (Section 4.1) | 130, 117, 92 | 113.0 | 2.7 / 6 | 0.56 |
| sign-only feedback | 122, 104, 104 | 110.0 | 3.0 / 6 | 0.50 |

*Table 3. Comparison of out-of-band feedback variants (3 repetitions each, budget 6). In both variants the mirage is 0 by definition and the deployment trajectory is monotone (Section 3). What is compared is the real-world output the gate reaches and the quality of the candidate stream.*

The result points to the gate as the load-bearing component. Even the sign-only gate, which never sees the oracle numbers, reached a mean final output of 110.0, similar to the 113.0 of full feedback (per-repetition ranges 104 to 122 versus 92 to 130, overlapping), and provided the same structural guarantee of preserving the best state reached. The regression rate of the candidate stream (wasted-cycle ratio) was also similar, 0.50 versus 0.56, so no evidence was observed at this scale that oracle-number feedback materially steers candidate generation. What remains as the source of the in-band versus out-of-band gap is therefore not the information content of the feedback but what the gate is grounded in, consistent with the conclusion of Section 4.3. Since this is a 3-repetition pilot, we do not claim the information effect is absent; establishing its magnitude is deferred to the expanded repetitions of the full measurement after the preregistration freeze.

### 4.7 Anatomy of the Progress Mirage

A single trajectory compresses the phenomenon (the first in-band-self repetition, straight from the raw log). In the first three cycles the agent made genuine improvements: removing two pre-friction steps from the funnel took conversions from 33 to 47 (+14); rewriting a passive landing page with a benefit-led headline and a clear CTA took them to 95 (+48); removing a verification step took them to 104 (+9). Up to this point the loop's self-record and the world agreed. Over the next three cycles the agent continued with conversion-optimization orthodoxy: social-proof copy above the CTA (-6), an inline email input on the landing page (-1), a named testimonial quote (-6). With each change the agent claimed it would raise conversion, and in-band-self accepted them all. The deployed state slid from 104 to 91. In the loop's self-record, six improvements accumulated; in the world, three steps forward were followed by three steps back.

This is the anatomy of the progress mirage. Even a capable frontier agent cannot tell, from inside the transcript, whether its change raised or lowered a real-world metric. In the early phase, the large levers lie where plausibility and effect coincide, so the self-verdict is also correct. Once the levers are exhausted, plausibility and effect diverge, and an evaluator trapped in the transcript cannot see the divergence. When the evaluator reads the world, the same regressions are rejected and the state reached is preserved.

### 4.8 The Motivating Field Observation

The phenomenon our measurement targets was observed in a real deployment before the controlled testbed. A production consumer web service operated an autonomous director-worker loop for about six weeks, 61 cycles. Completion of each cycle was judged by the loop itself, on the basis of the conversation transcript and analysis that cycle produced. That is, it was an in-band self-verdict, and the acceptance criterion was the production of a new analytical frame.

The consequence is isomorphic to our in-band-self arm. All 57 evaluated cycles were accepted, while over the same period every ground-truth signal outside the loop (arrival ledger, web analytics, append-only signup ledger, payment records, search indexing) was zero or flat. The loop's internal verdicts indicated progress; the state of the world indicated stagnation. More telling is that the loop did not stop even after recognizing the stagnation. Because the reward function rewarded analysis rather than outcomes, even awareness of stagnation could be accepted as part of a new analysis. When the reward function was moved to real-world outcomes and enforced by an out-of-band verifier, the behavior of the immediately following cycles reversed from analysis to shipping. This single observation is not a controlled experiment and cannot establish causality; we present it as motivation only. The causal evidence is provided by the controlled measurements of Sections 4.1 through 4.7.

### 4.9 Scope of the Results

The results of this section are a pilot: one open-ended task family and one boundary task, one fixed agent model, one stronger judge model, and 3 repetitions per arm. The measurements reported in this paper were performed with apparatus v2 — a strengthened judge receiving verdict history and diffs, per-run boundary proofs, and per-cycle canary scans — and the mirage ordering across arms was identical in earlier apparatus versions as well (apparatus history in the appendix). The direction of the effect is clear, but its magnitude is specific to this apparatus's feature weights (Section 6), and generalization to other domains, models, and longer horizons is to be settled by the full measurement after the preregistration freeze (multiple models, a second oracle, the reward-function axis, sensitivity sweeps).

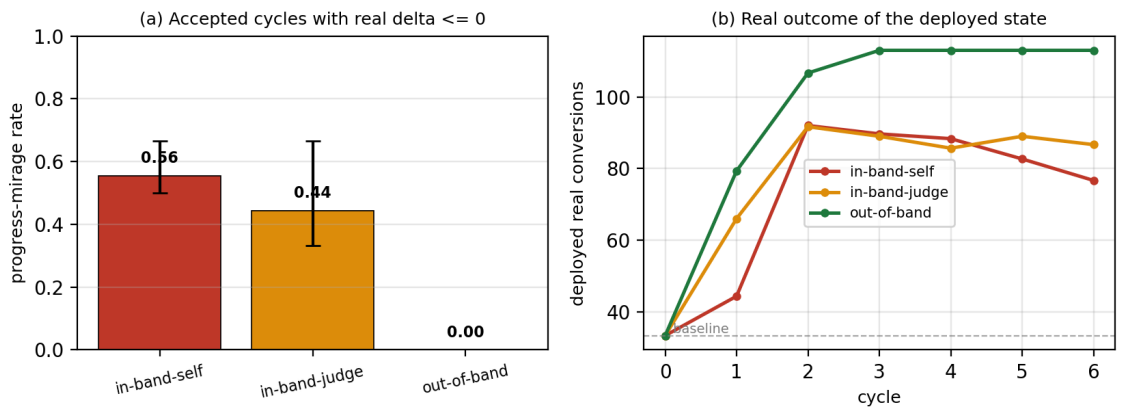

Figure 4. Pilot measurement results (a real frontier agent, isolated containers, 3 repetitions per arm, apparatus v2). (a) Progress-mirage rate by evaluator arm. Bars are repetition means; error bars are min and max. For in-band-self and the strengthened in-band-judge, a substantial fraction of accepted cycles in fact had an oracle delta of zero or below, whereas the out-of-band 0 is a definitional consequence of the gate rule. (b) Trajectories of actual conversions in the deployed (last-accepted) state (repetition means). All three arms rise on the first cycle's large lever; the self-verdict then keeps deploying plausible regressions and erodes from the peak, the judge filters some regressions so erosion is gentler, and out-of-band deploys no regressions and preserves the state reached.

## 5 Discussion

The measurements of Section 4 showed, under controlled conditions, that the progress mirage is not an accident of a single service but structurally inherent to long-running agent loops in which the generator evaluates its own completion. This section interprets those results. We first analyze why in-band evaluation is not sufficient (5.1), present the core principle of out-of-band evaluation (5.2), and lay out the auxiliary design principles that support it (5.3). We then generalize the relationship to the built-in agent primitives offered by contemporary platforms via the distinction of “loop plumbing vs. the reward brain” (5.4), and finally discuss when loop engineering is not justified — the conditions under which its costs exceed its benefits (5.5).

### 5.1 Why In-Band Evaluators Are Not Sufficient

The root of the problem lies in what the evaluator reads. An in-band evaluator issues ACCEPT/REJECT verdicts based on the transcript the generator produced, within the same information channel as the generator. In the measurements of Section 4, more than half of the cycles accepted by the two transcript-bound evaluators (the self-verdict and the strongest judge) were in fact real-world regressions, and both evaluators’ acceptance behavior was effectively independent of the sign of the oracle delta (Table 1 and the conditional acceptance rates of Section 4.3) — precisely because of this difference in channel.

For in-band evaluation to be sufficient, one premise must hold: the ground truth needed for the verdict must be contained in the transcript. But when the reward is grounded in real-world outcomes, this premise fails. How many users arrived, how many signups occurred, whether a payment happened, whether a page was indexed — these are not facts that exist inside the generator’s conversation. They exist in the state of the world: databases, version control, web analytics, append-only ledgers. The transcript can contain the generator’s narration of that world, but it cannot contain the world itself.

Self-evaluation bias arises in this gap. When the evaluator reads only the generator’s narration, the only signal available to it is how plausible the change is. And changes produced by a powerful LLM are almost always plausible. The archetype from Section 4: the strongest in-band-judge accepted the addition of star ratings and user counts as frictionless social proof, while actual conversion fell. The judge can assess plausibility against conversion-optimization orthodoxy, but what effect that orthodoxy actually has in the world is not in the transcript, so it cannot know. That all 57 cycles in the motivating field observation were accepted while real-world signals stayed at zero (Section 4.8) is likewise not a chance malfunction but a consequence of the same structure. An evaluator can only reward what it can see, and an evaluator that sees only the transcript can only see process.

Note that the in-band judge’s errors are not one-directional. The judge rejects real-world improvements for the same reason it accepts real-world regressions (Section 4.5). In a terrain where plausibility diverges from effect, there are genuine improvements that look implausible, and a transcript-only judge rejects them. The problem of in-band evaluation is thus not leniency but the absence of discrimination — which is why this paper reports rejection errors (false rejections) alongside acceptance errors (the mirage).

This is the lesson of HA1′s rejection. If the gap were a matter of judge quality, a stronger and more critical judge should close it. Yet in the measurements of Section 4, the strongest in-band-judge’s mirage rate was effectively the same as in-band-self’s. No matter how critical the judge is made, it cannot see ground truth that is not in the transcript. For open-ended real-world objectives, then, the limit of evaluation is lifted not by scaling the judge but by moving the evaluator’s information channel to the world. This is where our work departs from prior measurements that vary judge strength within the transcript channel, and it also suggests a boundary condition for the scalable-oversight family focused on building better judges (debate, recursive reward modeling, weak-to-strong): there exist objective domains where the ceiling is the judge's information channel, not the judge's quality.

Especially dangerous is that even when the generator recognizes stagnation, in-band evaluation cannot convert that recognition into a corrective signal (Section 4.8). Even awareness of stagnation is recorded in the transcript as part of a new analysis, and therefore becomes rewardable process too. As a result, the generator converges on measuring itself rather than producing outcomes — because measurement always generates new transcript, it is the most easily rewarded behavior under in-band reward. In short, in-band evaluation is sound only insofar as the reward is transcript-checkable; the moment the reward is grounded in the world, it becomes a device that mistakes process for outcome.

### 5.2 The Core Principle: An Out-of-Band Evaluator with Independent World Access

The core principle we propose is to separate the generator (agent) from the evaluator and to grant the evaluator independent world access rather than the transcript. Concretely, the evaluator has the following three properties.

First, the evaluator runs as a separately scheduled process. It is not an appendage invoked inside the generator's conversation turns but a distinct executing entity on a schedule independent of the generator. This separation is the heart of out-of-band. Only when evaluation happens outside the generator's context does it remain uncontaminated by the generator's narration.

Second, the evaluator does not trust the generator's self-judgment and does not read the transcript. It reads ground-truth signal sources directly: databases, version control, web analytics, append-only ledgers. Moving what the evaluator reads from the generator's output to the state of the world is the only structural means of blocking self-evaluation bias. When the evaluator looks at the world directly, no matter how plausible an analysis the generator produces, the world's zero reads as zero.

Third, the reward function must encode outcomes, not process. The evaluator's verdict rule must be not "was a new frame produced" but "did a real-world signal move, or is there at least verifiable evidence of traction that could move it (a commit or a live URL)." The out-of-band gate of the measurement apparatus (accept only when the oracle delta is positive) and the verifier of the field observation in Section 4.8 (reject if the sum of real-world signals is zero and there is no valid traction evidence) are two instances implementing this rule in its simplest form. Without this redefinition of the reward function, out-of-band access alone is meaningless: even if the evaluator reads the world, nothing changes if it uses those readings to reward process.

When these three properties operate together, the reward function finally becomes a real-world reward function. And crucially, the evaluator's reject is not a mere refusal: it is recorded as a rejection artifact, an input the next iteration is forced to read first. The rejection is thereby embedded in the starting condition of the next cycle, and the generator begins its next work confronting its own most recent failure. We interpret the shift from measuring to shipping in the cycles immediately after the intervention in the field observation of Section 4.8 as arising because the evaluator read the world directly and returned the result as the forced input of the next boot. To emphasize again, this is an interpretation of a single case, not a proof of causality. The causal evidence is provided by the measurements of Section 4, which implement the same feedback structure under controlled conditions.

### 5.3 Auxiliary Design Principles

If the out-of-band evaluator is the core principle, several auxiliary principles prop it up so that it operates reliably. One point must be made clear up front: these auxiliary principles are design notes derived from the field observation of Section 4.8, and our controlled measurements did not validate them individually. What the measurements support is the out-of-band gate and rejection feedback only; the principles below should be read as unvalidated prescriptions for making that gate run reliably in a real deployment.

**Append-only state.** If the loop is allowed to mutate its own state in place, that state readily becomes a "lie generator." When the generator overwrites the previous cycle's conclusions, neither the evaluator nor a human can reconstruct after the fact what actually happened. This is why ground-truth signal sources such as arrival ledgers and signup ledgers must be append-only. When only appends are allowed and overwrites are forbidden, past zeros remain past zeros, and the generator loses the room to re-narrate them as progress. For the out-of-band evaluator to obtain trustworthy signals, the signal sources must be immutable.

**File-based handoff.** Information transfer between generator and evaluator — above all the delivery of rejection artifacts — happens through external files, not the conversation context. When the evaluation's output persists as a file and the next boot is forced to read that file, the rejection signal does not depend on the volatility of the generator's context window. This is a simple mechanism that keeps evaluation and generation loosely coupled while guaranteeing that a rejection reaches the next cycle.

**Frame-adversarial review.** If the generator thinks only in directions that agree with its own assumptions, new frames become nothing more than tools for rationalizing self-evaluation bias more elaborately. Preventing this requires a review step that forces the generator to attack its own assumptions. In the field observation of Section 4.8, the operating rule that forced every cycle to read the real-world arrival ledger first and compute how many days the metrics had been flat (days-flat) can be seen as a frame-adversarial device that made the generator confront falsifiable facts before presenting its "things are going well" self-narrative.

**Self-healing scheduling.** The out-of-band evaluator must keep running independently of the generator, and stably — because the moment the evaluation process dies, the loop regresses to a self-grading state. The evaluator's scheduling must therefore be designed to recover on its own (self-healing) even when sessions drop or processes halt. The evaluator's availability is not an accessory feature; it is the integrity of the reward function itself.

### 5.4 Relation to Platform Built-in Primitives: Loop Plumbing vs. the Reward Brain

Contemporary agent platforms provide several built-in primitives for composing autonomous loops — representatively, a "loop" feature in which the model itself judges the result after every turn, and a "goal" feature in which the model evaluates whether the goal has been achieved. The evaluation style these primitives adopt is essentially in-band: when a turn ends, a single model reads that turn's transcript and judges completion.

We do not claim this in-band style is wrong. On the contrary, it has an appropriate scope of application. When the task is bounded and completion is transcript-checkable — whether output of a specific format was generated, whether all required items were mentioned, whether the code compiles — the answer is contained in the transcript. For such tasks, a model reading the transcript and judging is sufficient, and building a separate out-of-band evaluator is unnecessary overhead.

The problem is when the goal is an open-ended real-world objective. As discussed in Section 5.1, in that case the ground truth needed for the verdict is not in the transcript. Organic user growth, signups, payments, indexing are states of the world, and no matter how precisely one adjudicates the transcript, one cannot reach that world. For such goals, in-band evaluation is structurally insufficient, and out-of-band evaluation that reads the world directly is required. This is why our contribution complements platform primitives rather than replacing them.

Generalizing this distinction yields "loop plumbing vs. the reward brain" (Figure 3). Loop plumbing is the mechanical part that keeps an autonomous loop turning: the scheduler, pacing between cycles, session recovery. The reward brain is the part that decides what counts as progress: the domain-grounded evaluator. The central claim is this: loop plumbing can be delegated to platform primitives, but the reward brain cannot. Plumbing generalizes independent of domain, whereas the reward brain requires domain knowledge of which world signals constitute real progress for that service. The platform can turn the loop, but the platform cannot know what the loop should be turning toward.

### 5.5 Costs and Limits: When Loop Engineering Is Not Justified

Out-of-band evaluation is not free. Whether to adopt it depends on comparing costs and benefits, and in the following cases loop engineering is not justified.

**The cost of building and maintaining the out-of-band validator.** Creating a separate evaluation process, wiring up independent access to ground-truth signal sources, and keeping it running in a self-healing manner carry real engineering cost. If the work finishes quickly, is one-off in nature, or has transcript-checkable completion, this overhead exceeds the benefit. Loop engineering is justified for long-running autonomous loops — cases where evaluation can go cumulatively wrong and waste resources over a long period.

**The service-specificity of the reward function.** The reward brain is bound to its domain (Section 5.4). Which signals constitute real progress differs from service to service, so one service's reward function cannot be transplanted to another as-is. This means out-of-band evaluation is not a reusable generic component but a part that must be designed anew for each service. In early stages where the domain signal is unclear or not yet defined, what to reward is itself undecided, and fixing the reward function may be premature.

**The risk of over-constraint.** Narrowing the reward function to outcomes narrows the generator's action space by the same amount. There are two concrete instances. In the field observation of Section 4.8, the operating mode triggered after stagnation crossed its threshold blocked cycles that only analyzed and cycles that only touched policy wording. And the out-of-band gate of our measurement (accept only when the delta is positive), as seen in Section 4.5, eliminates the room for judgment that seems reasonable in real-world terms, and also rejects all valid steady-state edits with delta zero (refactoring, maintenance edits). This is the right prescription in a stagnation regime, but applying the same constraint indiscriminately can also block legitimately needed exploratory analysis or valley-crossing search (paths that accept temporary regress). Not all analysis is a symptom of self-evaluation bias; some analysis is genuine groundwork for the next action. If the reward function defines outcomes too narrowly and too short-term, the generator can be over-constrained into chasing only measurable short-term signals at the expense of long-term value. Designs that switch constraints on and off by situation — e.g., threshold-based mode switching that activates an outcome-forcing mode only when days of stagnation cross a threshold — are therefore important. The goal of out-of-band evaluation is not to bind the generator but to force the generator to face the fact when the world has stopped moving.

Finally, we restate the status of the evidence. The causal evidence of this work is the controlled pilot measurement of Section 4; the field observation of Section 4.8 is an anonymized single case providing motivation; and the auxiliary principles of Section 5.3 are unvalidated design notes derived from that field case. Whether the out-of-band evaluation architecture proposed here works identically in other domains, with other models, over longer time horizons, and whether each auxiliary principle is in fact a necessary condition, remain questions for the preregistered full measurement and follow-up research.

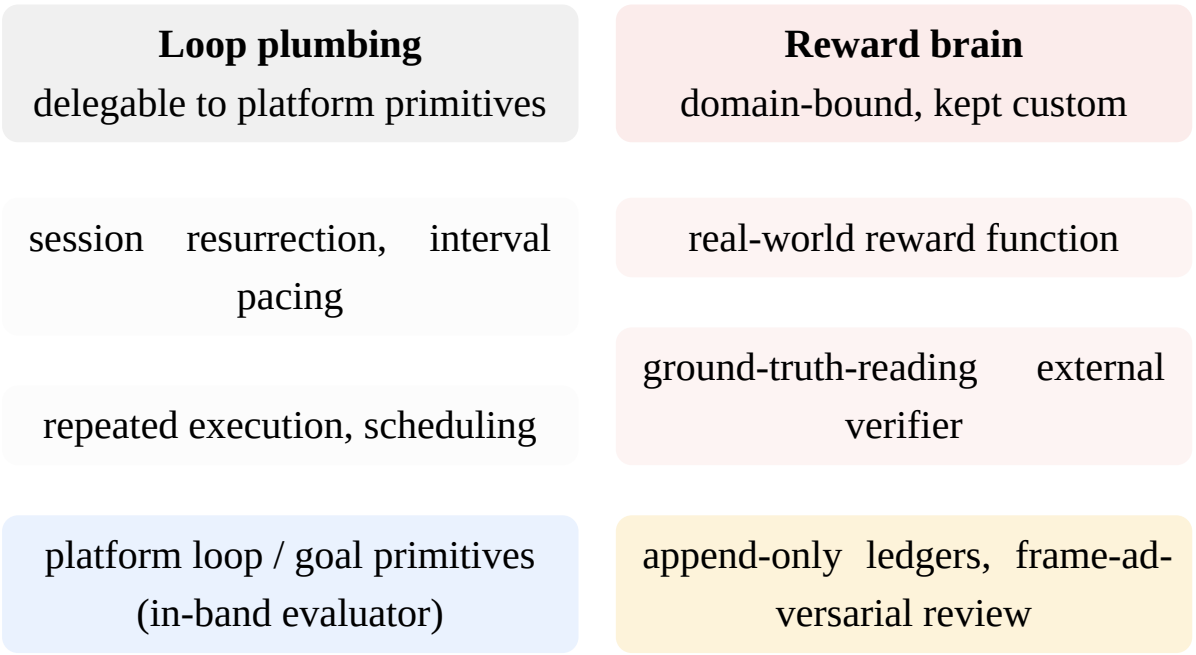

Figure 3. The distinction between loop plumbing and the reward brain. Plumbing such as scheduling, pacing, and session resurrection can be delegated to a platform's built-in loop/goal primitives, but the real-world reward function and the ground-truth-reading external verifier are domain-bound and cannot be delegated. The evaluator of the built-in primitives is in-band (it sees the conversation), whereas the open-ended real-world goals treated in this paper require out-of-band evaluation.

## 6 Threats to Validity

This section makes explicit the threats to the validity of this work. Because the core evidence is the controlled pilot measurement and the field observation serves only as motivation, the weight of the threats shifts from the limits of causal claims to the construct validity of the measurement apparatus and the scope of the pilot. We do not hide these threats; alongside each, we describe how the preregistered full measurement (PREREGISTRATION.md) is designed to resolve it further.

### 6.1 Construct Validity: Does the Oracle Capture Real Value?

This is the largest threat to this work. The measurement's ground truth is the conversion count produced by a synthetic user simulator — a weighted logistic over four properties of the deployed artifact (CTA presence, payload size, number of funnel steps, copy length). The foreseeable objection runs: if the simulator was designed so that plausibility and effect diverge, isn't the transcript judge's defeat a built-in conclusion?

Our defense is twofold. First, independence. The oracle's values are produced by an independent process from the physical properties of the deployed artifact, not from the agent's narrative, and the existence of the in-band vs. out-of-band gap derives from this independence, not from any particular weights. Second, the divergence between plausibility and effect is not an artifact of the simulator but a documented property of real-world conversion optimization. The large-scale online controlled experiment literature has repeatedly demonstrated that a majority of ideas experts judge plausible (around two-thirds by some reports) fail to move metrics or actively hurt them in real experiments [Kohavi et al., 2009], [Kohavi, Tang and Xu, 2020]. The very reason practitioners run A/B tests is that plausibility does not reliably predict effect. Our simulator reproduces that sign structure (a majority of plausible decorative edits being neutral or negative on the metric) in a simple form.

Even so, the existence of the gap and its magnitude must be distinguished. The absolute level of the mirage rate this pilot reports is specific to this simulator's feature weights; with different weights (e.g., giving social proof a positive channel), the agent's regression-generation rate and the magnitude of the mirage would also change. Our claim is confined to the existence and direction of the gap; we do not claim generalization of its magnitude. For reference, the mirage ordering across arms was invariant across the three versions of the measurement apparatus (before the gate rollback was introduced, before CTA-detection reinforcement, and final; apparatus history in the appendix). The preregistered full measurement includes a sensitivity sweep over feature weights (including a variant that puts the correlation between plausibility and effect on a dial) and a second oracle type (server-side event logs) that reduces dependence on the simulator.

Finally, we state the limit that the oracle is a measurement channel, not a value function. Because this oracle counts only conversions, it rewards as a conversion increase even a change that, say, removes a confirmation step and inflates unverified signups. Indeed the log contains a case in which the judge rejected such a change for a reason sound in real-world terms (unverified-signup inflation), yet by the oracle's standard it was tallied as a false rejection of a real improvement (Section 4.5). That out-of-band evaluation tied to a narrow metric can reproduce the surrogate-metric problem one level up is an honest limit of this design; in real deployments, the construct validity of the oracle metric itself (quality-adjusted metrics, plural signals) becomes a separate design problem.

### 6.2 Internal Validity: Is the Evaluator Channel Really the Only Difference?

The pilot holds the agent model, tool surface, task, and cycle budget identical across all arms and swaps only the evaluator gating the loop, so the measured effect is attributable to the evaluator's information-channel type. The rollback rule, under which rejected cycles are not promoted to deployment, makes the gate actually govern state, causally linking the real-outcome measurements to the evaluator. Two confounds remain. First, agent stochasticity: each arm's trajectory is an independent agent run, so the arms do not share identical rollouts; this is mitigated by multiple repetitions. Second, the information content of the feedback: each arm's gate verdict returns to the loop as a different feedback string, and in particular the out-of-band feedback contains the post-gate oracle scalars (the delta and current value). The out-of-band treatment is thus not a pure gate arrangement but the combination of a gate and ground-truth feedback, and indeed a second-order effect is observed in which the agent's regression-candidate generation rate falls in the out-of-band arm. This combination is a deliberate treatment that faithfully models what a real deployment's out-of-band evaluator actually provides (rejection rationales). This pilot additionally ran a sign-only variant that returns only the sign, providing first data for this decomposition (Section 4.6): a gate that never sees the oracle numbers largely preserved real-world output, so the gate was observed to be the load-bearing component; but at a scale of 3 repetitions, the precise magnitude of the information effect is to be settled by the expanded repetitions of the preregistered full measurement.

### 6.3 Limits of Ungameability

Oracle isolation is proven by measurement, not assumed. In the measurements reported here, a boundary proof was performed at the start of every run (each combination of arm and repetition), agent-side output and artifacts were scanned every cycle for the oracle canary string, and the apparatus was wired to abort measurement on any canary occurrence (zero occurrences). But this proof covers network paths and filesystem access; it does not rule out every other bypass. We state the remaining surfaces. (a) File-layer attacks such as symlinks through the shared artifact volume (this can be narrowed by real-path verification on the oracle-side read). (b) Because the in-band-judge input includes the raw text of files the agent wrote, an adversarial agent could plant verdict instructions in a file — prompt injection (not observed in this measurement, and under our threat model the agent was not induced to attack the evaluator). (c) Because the out-of-band feedback exposes one oracle scalar per cycle, over a very long horizon the feature weights could be reverse-engineered from numeric feedback alone. That is, the apparatus's guarantee is unfakeability (the agent cannot manufacture oracle values by narrative), not long-horizon unoptimizability. Also, while the agent does not know the simulator's feature set and weights in advance, the narrowness of a four-feature surface is a simplification that is safe only at pilot scale. The full measurement narrows this surface with feature expansion and adversarial red-team scenarios.

### 6.4 External Validity: The Scope of the Pilot

This measurement is a pilot with one open-ended task family, one fixed agent model, one stronger judge model, and 3 repetitions per arm. Whether the same effect holds in other domains, with less capable models, and over longer horizons is beyond this pilot's scope. The synthetic user is an approximation of real users; in particular, this oracle responds instantly and without noise. Real-world metrics are noisy and delayed, and under those conditions a simple gate that accepts only positive deltas can reject true improvements for lack of statistical power. Gate design under noisy oracles (e.g., interval-estimate-based acceptance) is an extension axis of the full measurement. Full resolution of external validity is deferred to replication on a real service.

### 6.5 Conclusion Validity and Prevention of Selective Reporting

The per-cell repetition count of the pilot is small. The intervals in Table 1 are bootstrap percentile intervals over 3 trajectories — at this sample size, closer to observed ranges than calibrated 95 percent confidence intervals. Definitive conclusions about effect size and durability are deferred to the full measurement after the preregistration freeze. Analysis is performed by standard-library scripts over raw cycle logs with no human discretion, and every registered dependent variable is reported even when the result is unfavorable to our hypotheses or uninformative (e.g., cycles to first real outcome identical across arms) (Section 4.1). Definitional differences between the registration document and the paper are not concealed but disclosed as amendments, with numbers reported under both definitions (Section 4.1).

### 6.6 Reproducibility

The analysis is fully reproducible: applying the analysis scripts to the released raw cycle logs recomputes every number in the paper. The measurement trajectories themselves include non-deterministic LLM calls and so differ across executions; what is fixed is the simulator's cohort randomness. Within a repetition the same seed is used across cycles (common random numbers), so the conversion count is a monotone deterministic function of artifact features and the sign of a delta cannot be flipped by sampling noise. The testbed is released together with its container configuration; the field observation used as motivation is anonymized, and the original deployment itself is not reproducible.

### 6.7 Cost and Over-Constraint Risk

Out-of-band evaluation carries cost: additional oracle queries and infrastructure. In the testbed, oracle queries are cheap, but the possibility that in real deployments the cost offsets the benefit (adversarial hypothesis HA2) was not tested in this pilot and is tested by the budget-normalized analysis of the preregistered full measurement. Moreover, if the reward function is tied only to outcome metrics, it can over-constrain the agent's exploration and miss better solutions (Section 5.5). This over-constraint risk is mitigated by designs that tie the reward function to outcomes while leaving paths open, and by threshold-based mode switching, but the right point of balance is beyond the scope of this work.

## 7 Conclusion

When a long-running autonomous agent loop grades its own work, self-evaluation bias makes it mistake analysis for progress: the loop reports advancement while real-world outcomes stagnate. This paper named this failure mode the progress mirage and diagnosed its cause as the in-band structure that traps the evaluator inside the conversation transcript.

As the remedy, we proposed an architecture that separates the generator from the evaluator and places the evaluator out-of-band. The evaluator runs as a separate process, accesses real-world ground truth directly rather than the conversation, enforces a real-world reward function, and, when the function is not satisfied, leaves a rejection artifact that the next iteration is forced to confront. Append-only state, file-based handoff, frame-adversarial review, and a self-healing scheduler were presented alongside as auxiliary design notes derived from field observation; their individual validation is future work.

We tested this diagnosis and remedy with a controlled pilot measurement. In a testbed that holds the agent and tool surface fixed and swaps only the information-channel type of the evaluator gating the loop, we enforced a world-state oracle that is ungameable in principle via container and network isolation and proved that isolation by per-run measurement. Running a real frontier agent repeatedly under three evaluator arrangements, the in-band self-verdict that sees only the transcript accepted as progress a substantial fraction of cycles in which real-world metrics stagnated or regressed, whereas the out-of-band evalua-tion that queries the world-state oracle did not; and even the strongest in-band judge, reading the transcript critically, failed to close the gap. The cause of the gap is not the evaluator's quality but what the evaluator is grounded in.

The practical implication is summarized by the distinction between loop plumbing and the reward brain. Plumbing — scheduling, pacing, session revival — can be delegated to the platform's built-in loop/goal primitives; the reward brain grounded in the real world — the out-of-band evaluator with access to ground truth and the domain reward function — cannot. Because the evaluator of the built-in primitives is in-band, looking at the conversation, it suffices for well-bounded transcript-verifiable tasks but is structurally insufficient for open-ended real-world objectives whose success signal lies outside the transcript.

This measurement is a pilot on one task family and one agent model, and the motivating field observation is an anonymized single-deployment observation, not causal proof. Future work is to freeze the preregistration (PREREGISTRATION.md) and then extend the measurement with multiple models, two kinds of world-state oracle, and a transcript-verifiable contrast task testing the boundary hypothesis, to settle the generalization scope and durability of the effect, and to quantify the cost-benefit of out-of-band evaluation. Nevertheless, this pilot puts one fact on the record by measurement. As we attach autonomous agents to ever more open-ended real-world goals, grounding the evaluator in the world rather than the transcript is not a choice but a structural requirement for preventing the progress mirage.

## References

[Advani 2026] L. Advani, "From Confident Closing to Silent Failure: Characterizing False Success in LLM Agents," arXiv:2606.09863, 2026.

[Amodei et al. 2016] D. Amodei, C. Olah, J. Steinhardt, P. Christiano, J. Schulman, and D. Mané, "Concrete Problems in AI Safety," arXiv:1606.06565, 2016.

[Anthropic 2024] Anthropic, "Building Effective AI Agents," Anthropic Engineering Blog, 2024.

[Anthropic 2025] Anthropic, "Effective Harnesses for Long-Running Agents," Anthropic Engineering Blog, 2025.

[Anthropic 2025b] Anthropic, "Agent Harness Design Notes," Anthropic Engineering Blog, 2025.

[Atinafu and Cohen 2026] Y. Atinafu and R. Cohen, "Reward-HackingAgents: Benchmarking Evaluation Integrity for LLM ML-Engineering Agents," arXiv:2603.11337, 2026.

[Bai et al. 2022] Y. Bai et al., "Constitutional AI: Harmlessness from AI Feedback," arXiv:2212.08073, 2022.

[Baker et al. 2025] B. Baker, J. Huizinga, L. Gao, Z. Dou, M. Y. Guan, A. Madry, W. Zaremba, J. Pachocki, and D. Farhi, "Monitoring Reasoning Models for Misbehavior and the Risks of Promoting Obfuscation," arXiv:2503.11926, 2025.

[Burns et al. 2023] C. Burns et al., "Weak-to-Strong Generalization: Eliciting Strong Capabilities with Weak Supervision," arXiv:2312.09390, 2023.

[Çağatan and Zhao 2026] Ö. V. Çağatan and X. Zhao, "Reward Hacking in Language Model Agents: Revisiting AI Safety Gridworlds," arXiv:2606.15385, 2026.

[Caldwell et al. 2026] S. Caldwell, M. Harley, A. Dawson, M. Kouremetis, V. Abruzzo, and W. Pearce, "ScopeJudge: Cost-Aware Pre-Execution Gating for Offensive Security Agents," arXiv:2607.07774, 2026.

[Christiano et al. 2017] P. Christiano, J. Leike, T. B. Brown, M. Martic, S. Legg, and D. Amodei, "Deep Reinforcement Learning from Human Preferences," in *Advances in Neural Information Processing Systems (NeurIPS)*, 2017.

[Flynt 2026] J. Flynt, "GroundEval: A Deterministic Replacement for LLM-as-Judge in Stateful Agent Evaluation," arXiv:2606.22737, 2026.

[Gabor et al. 2025] J. Gabor, J. Lynch, and J. Rosenfeld, "EvilGenie: A Reward Hacking Benchmark," arXiv:2511.21654, 2025.

[Gao et al. 2023] L. Gao, J. Schulman, and J. Hilton, "Scaling Laws for Reward Model Overoptimization," in *Proc. Int. Conf. Machine Learning (ICML)*, 2023.

[Huang et al. 2024] J. Huang et al., "Large Language Models Cannot Self-Correct Reasoning Yet," in *Proc. Int. Conf. Learning Representations (ICLR)*, 2024.

[Irving et al. 2018] G. Irving, P. Christiano, and D. Amodei, "AI Safety via Debate," arXiv:1805.00899, 2018.

[Jimenez et al. 2024] C. E. Jimenez et al., "SWE-bench: Can Language Models Resolve Real-World GitHub Issues?," in *Proc. Int. Conf. Learning Representations (ICLR)*, 2024.

[Kohavi et al. 2009] R. Kohavi, R. Longbotham, D. Sommerfield, and R. M. Henne, "Controlled Experiments on the Web: Survey and Practical Guide," *Data Mining and Knowledge Discovery*, vol. 18, no. 1, pp. 140–181, 2009.

[Kohavi, Tang and Xu 2020] R. Kohavi, D. Tang, and Y. Xu, *Trustworthy Online Controlled Experiments: A Practical Guide to A/B Testing*. Cambridge University Press, 2020.

[Krakovna et al. 2020] V. Krakovna et al., "Specification Gaming: The Flip Side of AI Ingenuity," DeepMind, 2020.

[Leike et al. 2018] J. Leike, D. Krueger, T. Everitt, M. Martic, V. Maini, and S. Legg, "Scalable Agent Alignment via Reward Modeling: A Research Direction," arXiv:1811.07871, 2018.

[Li et al. 2026] X. Li, W. Yan, Y. Wu, P. Liang, M. Yuan, J. Liu, and J. Yang, "Early Diagnosis of Wasted Computation in Multi-Agent LLM Systems via Failure-Aware Observability," arXiv:2606.01365, 2026.

[Lightman et al. 2023] H. Lightman et al., "Let's Verify Step by Step," in *Proc. Int. Conf. Learning Representations (ICLR)*, 2024 (arXiv 2023).

[Lù et al. 2025] X. H. Lù et al., "AgentRewardBench: Evaluating Automatic Evaluations of Web Agent Trajectories," arXiv:2504.08942, 2025.

[Ma et al. 2026] X. Ma, C. Zheng, J. Qiu, J. Hong, Y. Yao, X. Qu, J. Yin, X. Lou, J. Wang, W. Liu, W. Zhang, Z. Zhang, and H. Zhao, "Retrospective Progress-Aware Self-Refinement for LLM Agent Training," arXiv:2606.14302, 2026.

[Madaan et al. 2023] A. Madaan et al., "Self-Refine: Iterative Refinement with Self-Feedback," in *Advances in Neural Information Processing Systems (NeurIPS)*, 2023.

[Manheim and Garrabrant 2018] D. Manheim and S. Garrabrant, "Categorizing Variants of Goodhart's Law," arXiv:1803.04585, 2018.

[Ouyang et al. 2022] L. Ouyang et al., "Training Language Models to Follow Instructions with Human Feedback," in *Advances in Neural Information Processing Systems (NeurIPS)*, 2022.

[Pan et al. 2022] A. Pan, K. Bhatia, and J. Steinhardt, "The Effects of Reward Misspecification: Mapping and Mitigating Misaligned Models," in *Proc. Int. Conf. Learning Representations (ICLR)*, 2022.

[Pan et al. 2024] J. Pan, H. He, S. R. Bowman, and S. Feng, "Spontaneous Reward Hacking in Iterative Self-Refinement," arXiv:2407.04549, 2024.

[Panickssery et al. 2024] A. Panickssery, S. R. Bowman, and S. Feng, "LLM Evaluators Recognize and Favor Their Own Generations," in *Advances in Neural Information Processing Systems (NeurIPS)*, 2024.

[Shinn et al. 2023] N. Shinn, F. Cassano, E. Berman, A. Gopinath, K. Narasimhan, and S. Yao, "Reflexion: Language Agents with Verbal Reinforcement Learning," in *Advances in Neural Information Processing Systems (NeurIPS)*, 2023.

[Skalse et al. 2022] J. Skalse, N. H. R. Howe, D. Krasheninnikov, and D. Krueger, "Defining and Characterizing Reward Hacking," in *Advances in Neural Information Processing Systems (NeurIPS)*, 2022.

[Uesato et al. 2022] J. Uesato et al., "Solving Math Word Problems with Process- and Outcome-Based Feedback," arXiv:2211.14275, 2022.

[Wang et al. 2026] B. Wang et al., "The Verification Horizon: No Silver Bullet for Coding Agent Rewards," arXiv:2606.26300, 2026.

[Xie 2026] H. Xie, "Forage V2: Knowledge Evolution and Transfer in Autonomous Agent Organizations," arXiv:2604.19837, 2026.

[Yang et al. 2024] J. Yang et al., "SWE-agent: Agent-Computer Interfaces Enable Automated Software Engineering," in *Advances in Neural Information Processing Systems (NeurIPS)*, 2024.

[Yao et al. 2023] S. Yao, J. Zhao, D. Yu, N. Du, I. Shafran, K. Narasimhan, and Y. Cao, "ReAct: Synergizing Reasoning and Acting in Language Models," in *Proc. Int. Conf. Learning Representations (ICLR)*, 2023.

[Yao et al. 2024] S. Yao, N. Shinn, P. Razavi, and K. Narasimhan, "τ-bench: A Benchmark for Tool-Agent-User Interaction in Real-World Domains," arXiv:2406.12045, 2024.

[Zhao et al. 2026] B. Zhao, D. Srikanth, Y. Wu, and Z. Jiang, "SpecBench: Measuring Reward Hacking in Long-Horizon Coding Agents," arXiv:2605.21384, 2026.

[Zheng et al. 2023] L. Zheng et al., "Judging LLM-as-a-Judge with MT-Bench and Chatbot Arena," in *Advances in Neural Information Processing Systems (NeurIPS) Datasets and Benchmarks*, 2023.

Industry/engineering sources (agent harness design, etc.) will carry URLs and access dates in the submission version. Citations will be converted to the target venue's bibliography format at submission time.